\documentclass[10pt,twocolumn,letterpaper]{article}

\usepackage{cvpr}
\usepackage{times}
\usepackage{epsfig}
\usepackage{graphicx}
\usepackage{amsmath}
\usepackage{amssymb}
\usepackage{mathrsfs}
\usepackage{caption}
\usepackage{subcaption}

\usepackage[breaklinks=true,bookmarks=false]{hyperref}

\cvprfinalcopy 

\def\cvprPaperID{****} 

\begin{document}

\title{Universal, transferable and targeted adversarial attacks}

\author{Junde Wu, Rao Fu\\
Harbin Institute of Technology\\
{\tt\small wujunde@hit.edu.cn}
}

\maketitle

\begin{abstract}
Deep Neural Networks have been found vulnerable recently. A kind of well-designed inputs, which called adversarial examples, can lead the networks to make incorrect predictions. Depending on the different scenarios, goals and capabilities, the difficulties of the attacks are different. For example, a targeted attack is more difficult than a non-targeted attack, a universal attack is more difficult than a non-universal attack, a transferable attack is more difficult than a nontransferable one. The question is: Is there exist an attack that can meet all these requirements? In this paper, we answer this question by producing a kind of attacks under these conditions. We learn a universal mapping to map the sources to the adversarial examples. These examples can fool classification networks to classify all of them into one targeted class, and also have strong transferability. Our code is released at: xxxxx.
\end{abstract}

\section{Introduction}
Deep Neural Networks have outperformed many previous techniques in wide fields. Their high accuracy and fast speed make them to be widely used in real applications. Despite these great successes, they have been found vulnerable to the adversarial examples: the output of the networks can be manipulated by adding a kind of meticulously crafted subtle perturbations to the input data. This property is shown to be generally existing. Whether in the tasks of computer vision, like classification \cite{szegedy2013intriguing}, objection detection \cite{xie2017adversarial}, semantic segmentation \cite{fischer2017adversarial} or in tasks of Natural Language Processing \cite{jia2017adversarial} and Reinforcement Learning \cite{huang2017adversarial}.

\begin{figure}
     \centering
         \includegraphics[width=0.5\textwidth]{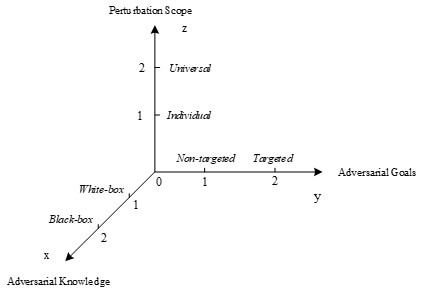}
        \caption{The taxonomy of adversarial attacks}
        \label{fig:taxonomy}
\end{figure}

In the classification task, the adversarial attacks aim to manipulate the network to misclassify. Previous works have proven that optimizing this kind of adversarial examples can be very cheap and effective \cite{goodfellow2014explaining}. But the difficulty of attacking varies with the adversarial goals, perturbation scope and adversary knowledge. To put it clearly, we taxonomize the threat models by different goals, adversary’s knowledge and perturbation scope. The taxonomy is shown in Figure \ref{fig:taxonomy}.

$\bullet$ Adversarial Goals

- \textit{Non-targeted misclassification} forces the victim model to incorrectly classify the source image into an arbitrary class.

- \textit{Targeted misclassification} forces the victim model to incorrectly classify all of the source images as a specific targeted class.

$\bullet$ Adversarial Knowledge

- \textit{White-box attacks} assume threat model knows everything about the victim model, including the network architecture and the training dataset. 

- \textit{Black-box attacks} assume threat model can not get access to the victim model. It only knows the standard output of the network, like the labels of the source images and the corresponding scores. But if the adversarial examples are transferable, a white-box attack can be transferred to a black-box model. 

$\bullet$ Perturbation Scope

- \textit{Individual attacks} solve the optimization problem for each single source image. The perturbations for each source image are all different.  

- \textit{Universal attacks} denote the attacks which are able to learn a universal mapping relation between the source images and the adversarial examples.

In Figure \ref{fig:taxonomy}, the difficulty of attacking increases with the axes $x,y$ and $z$. In this paper, we mainly explore how to produce attacks under the strictest conditions, which is correspond to the point (2,2,2) in Figure \ref{fig:taxonomy}, denoting the transferable, universal and targeted attacks.

The remainder of the paper is organised as follows. In Sec. 2 we briefly review methods proposed to generate adversarial examples. In Sec. 3 we propose low-frequency fooling images. In Sec. 4 we present our method of producing adversarial examples. In Sec. 5 we report the results of ablation experiments and contrast experiments. In Sec. 6 we discuss the experiment results and provide some reasonable explanations. Finally, in Sec. 7 we conclude our main contributions of the paper and propose several research lines that can be explored.

\section{Related work}
 All the attacks can be divided into white-box attacks and black-box attacks. Some methods can be directly deployed in a black-box attack since they don not require gradients. For example, Chen \textit{et al}. \cite{chen2017zoo} used symmetric difference gradient to estimate the gradient and Hessian matrix. Su \textit{et al}. \cite{su2019one} utilized differential evolution to find the optimal solution. Zhao \textit{et al}. \cite{zhao2017generating} built a generator to map the latent vector to the adversarial examples, and used search algorithms to search the effective noise. However, all these methods have to compromise on adversarial goals or perturbation scope. \cite{chen2017zoo} has to optimize on every single image. \cite{su2019one} aimed to generate adversarial examples by only modifying one pixel and only experimenting on small images. \cite{zhao2017generating} is non-targeted attack.

Most of other methods have to get access to the victim model, but due to the transferability of adversarial examples proposed by Papernot \textit{et al}.\cite{papernot2016transferability}, they are able to be transferred to other victim models \cite{rozsa2016adversarial}, \cite{goodfellow2014explaining}, \cite{papernot2016limitations}, \cite{moosavi2016deepfool}, \cite{carlini2017towards} or even the black-box server \cite{liu2016delving}. However, it is still hard to ensure good transferability when considering the distortion of adversarial examples, time consumption and targeted attack. Szegedy \textit{et al}. \cite{szegedy2013intriguing} first used L-BFGS method to generate the adversarial examples, but it is time-consuming. Lan \textit{et al}. \cite{goodfellow2014explaining} and the extended methods \cite{rozsa2016adversarial}, \cite{kurakin2016adversarial}, \cite{dong2018boosting}, \cite{tramer2017ensemble} performed only one step gradient at each pixel to speed up the optimization. Papernot \textit{et al}. \cite{papernot2016limitations} computed Jacobian matrix of given samples, tried to make most significant variances with smallest perturbations. Moosavi \textit{et al}. \cite{moosavi2016deepfool} further reduced the intensity of perturbation by considering the classifier is linearized around the samples. Carlini \textit{et al}. \cite{carlini2017towards} defined a new objective function to describe the distance between sources and adversarial examples for better optimizing the distance and penalizing term. Liu \textit{et al}. \cite{liu2016delving} optimized on ensemble deep neural networks to improve transferability. They also discovered that targeted attacks are harder to transfer than non-targeted attacks. However, all these optimization-based methods are individual attacks, which means they have to optimize on each single source image. Moosavi \textit{et al}. \cite{moosavi2017universal} first found the universal perturbation over large dataset, but this method is unable to be deployed to targeted attacks. In this paper, we first show the existence of the universal, transferable and targeted adversarial examples.

\section{Low-frequency fooling image}
Before going into the adversarial examples, let us discuss about the fooling images first. This nomenclature is adapted from \cite{nguyen2015deep}, which means the images that are meaningless to humans, but the networks classify them into certain classes with high confidences.

\begin{figure}
     \centering
     \begin{subfigure}[b]{0.2\textwidth}
         \centering
         \includegraphics[width=\textwidth]{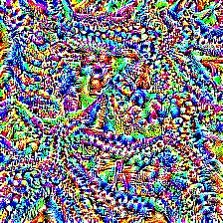}
         \caption{fooling image}
         \label{fig:fooling image}
     \end{subfigure}
     \hfill
     \begin{subfigure}[b]{0.2\textwidth}
         \centering
         \includegraphics[width=\textwidth]{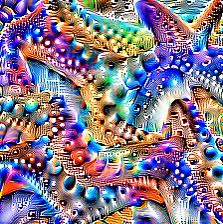}
         \caption{visualized image}
         \label{fig:visualized image}
     \end{subfigure}
        \caption{The comparison of fooling image and visualized image of class 'starfish' in VGG19. Both images maximize the activation of the last fully connected layer before softmax.}
        \label{fig:comparison with fooling}
\end{figure}

For producing a fooling image, we solve the following optimization problem: 
\begin{equation}\label{equation:fooling}
I_{f} = \mathop{\arg\max} P(y|I_{f})
\end{equation}
where $I_{f}$ denotes the fooling image, $P(y|I_{f})$ denotes the classifier confidence of the targeted label $y$ when inputting image $I_{f}$. If the neural networks are differentiable with respect to their inputs, we can use the derivatives to iteratively tweak the input towards the goal. The way to produce fooling images just like the way to visualize the network \cite{olah2017feature}. The difference between them is the target layer and constraint conditions. Neural network visualization optimizes the layers which it aims to visualize. And for recognition, neural network visualization will add extra constraints to this optimization problem, forcing the goal to lie in the low-frequency space. The contrast of fooling image and network visualization result is shown in Figure \ref{fig:comparison with fooling}. 
A natural question after the comparison is why the networks will naturally produce the high-frequency unrecognizable noises? Odena \textit{et al}. \cite{odena2016deconvolution} indicated these noises may be closely related to the structure of the networks, especially the strided deconvolutional layers and the pooling operations. Since Odena \textit{et al}. \cite{odena2016deconvolution}] pointed out the deconvolution operations are the root of the grid effect, one possible interpretation is when we leverage the gradients going backward from the targeted label, as what we do when solving Eqn. (\ref{equation:fooling}), every convolution layer in the network will serve as a deconvolutional layer. Therefore, the gradients will have to go through too many deconvolutional layers (generally 2 or 3 deconvolutional layers are able to produce grid effect). The grid effect is sequentially magnified by these deconvolutional layers, and finally become the high-frequency noises.

If these high-frequency noises are closely related to the structure of the networks, a plausible assumption is that the low-frequency fooling images will be unrelated to the structure of networks, thus being more general and transferable than these high-frequency ones. To test this hypothesis, we tried several methods as follows to constrain the high-frequency gradients when optimizing fooling images. 

1. Transformation Robustness (TR) constrains high-frequency gradients by applying small transformations to the fooling images before optimization. Here, we rotate, scale and jitter the images. The constrained optimization process can be expressed as:
\begin{equation}\label{equation:TR}
I_{f}^{tr} = \mathop{\arg\max} P(y|\ T(I_{f}^{tr}))
\end{equation}
where $T$ denotes the composition of the specific transformations.

2. Decorrelation (DR) decorrelated the relationship between the neighbour pixels. Here, we do it by using gradient descent in the Fourier basis, as what \cite{olah2017feature} did to visualize the network. It can be expressed as:
\begin{equation}\label{equation: DR}
\begin{split}
&\theta = \mathop{\arg\max} P(y|\mathcal{F}(\theta))\\
&I_{f}^{dr} = \mathcal{F}(\theta)
\end{split}
\end{equation}
where $\mathcal{F}$ denotes Fourier transform.

3. Transformation Robustness and Decorrelation (TR and DR) are able to combine together to generate fooling images, which is expressed as:
\begin{equation}\label{equation:TR}
\begin{split}
&\theta = \mathop{\arg\max} P(y|\ T(\mathcal{F}(\theta)))\\
&I_{f}^{tr\&dr} = \mathcal{F}(\theta)
\end{split}
\end{equation}

4. Gradient optimized Compositional Pattern Producing Network (Gradient-CPPN) uses CPPN \cite{stanley2007compositional} to produce fooling images. CPPN is a neural network that map a position of the image to its color. Thus, the frequency of the outputted fooling image is only related to the architecture of CPPN. The simpler the structure of the network, the lower the frequency of the images. This method optimizes CPPN parameters by the gradients of the victim model, which can be expressed as:
\begin{equation}\label{equation:G-CPPN}
\begin{split}
&CPPN(M_{p}) = \mathop{\arg\max} P(y|\ CPPN(M_{p}))\\
&I_{f}^{gcppn} = CPPN(M_{p})
\end{split}
\end{equation}
where $M_{p}$ is a 2-D position map.

5. CPPN encoded Evolutionary Algorithms (EA-CPPN) is proposed by \cite{nguyen2015deep}. This method uses CPPN encoded image to represent genomes and uses EA to optimize.  

\begin{figure}
    \begin{subfigure}[b]{0.2\textwidth}
         \centering
         \includegraphics[width=\textwidth]{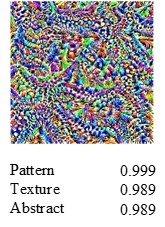}
         \caption{naive}
         \label{fig:fig2_naive}
     \end{subfigure}
     \hfill
     \begin{subfigure}[b]{0.2\textwidth}
         \centering
         \includegraphics[width=\textwidth]{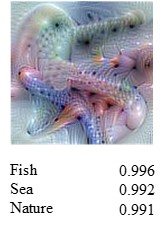}
         \caption{TR}
         \label{fig:TR}
     \end{subfigure}
     \hfill
     \begin{subfigure}[b]{0.2\textwidth}
         \centering
         \includegraphics[width=\textwidth]{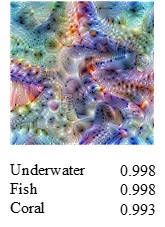}
         \caption{DR}
         \label{fig:DR}
     \end{subfigure}
     \hfill
     \begin{subfigure}[b]{0.2\textwidth}
         \centering
         \includegraphics[width=\textwidth]{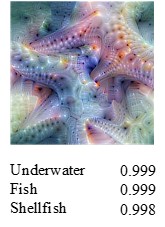}
         \caption{TR \& DR}
         \label{fig:TR and DR}
     \end{subfigure}
          \hfill
     \begin{subfigure}[b]{0.2\textwidth}
         \centering
         \includegraphics[width=\textwidth]{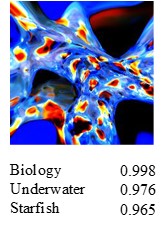}
         \caption{Gradient-CPPN}
         \label{fig:Gradient-CPPN}
     \end{subfigure}
          \hfill
     \begin{subfigure}[b]{0.2\textwidth}
         \centering
         \includegraphics[width=\textwidth]{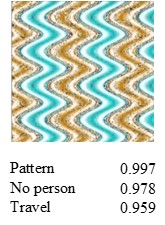}
         \caption{EA-CPPN}
         \label{fig:EA-CPPN}
     \end{subfigure}
        \caption{Some samples of high-frequency fooling image: \ref{fig:fig2_naive} and low-frequency fooling images generated by different methods: \ref{fig:TR}-\ref{fig:EA-CPPN}. The classes following the images are predicted by Clarifai.com.}
        \label{fig:comparison with low and high}
\end{figure}

We choose VGG19 \cite{simonyan2014very} as our victim model to train the fooling images and test the results on Clarifai.com, which is a black-box image classification server. We show some examples in Figure \ref{fig:comparison with low and high}. We find that in consistent with our hypothesis, low-frequency images can fool Clarifai.com to classify them as targeted or related classes, while high-frequency noises are fail to fool the system. In all these low-frequency images, gradient optimized CPPN performs better than the other methods.
These low-frequency fooling images are targeted and transferable, but may not be the adversarial examples. In the next section, we will introduce how to leverage these low-frequency images to produce the adversarial examples we want. For convenience, we refer to these constrained low-frequency fooling images as $I_{lf}$, and unconstrained high-frequency fooling image as $I_{hf}$.

\section{Universal, transferable, targeted attacks}
\subsection{Method} For producing the adversarial examples, we aim at mapping the distribution of the source images to the targeted adversarial distribution. The samples in this distribution should maintain the similarity with the source images in the low level (pixel level), while have the similar high-level features with $I_{lf}$. A reasonable assumption is that these high-level features preserve the attributions of $I_{lf}$: targeted and transferable. We prove this assumption by comparing the high-level features of fooling images with different frequencies. We find that the mean and variance of different distributions cluster together, which denotes that $I_{lf}$ have some specific properties to ensure their transferability. 

We build a conditional image generation function to shift the original source image distribution to inherit $I_{lf}$ distribution's properties. Note $q(I_{s})$ and $q(I_{lf})$ are the distributions of source images and low-frequency fooling images, and $I_{a}$ are the targeted adversarial examples. Our goal is to learn the conditional distribution $q(I_{s-lf}|I_{lf})$ to satisfy:
\begin{equation}\label{equation:adversarial}
I_{a} = \mathop{\arg\max}_{I_{a}\backsim q(I_{s-lf}|I_{lf})} P(y|I_{a})
\end{equation}
where $I_{s-lf}$ denotes the samples produced from $I_{s}$ to have the properties of $I_{lf}$.

We build an an encoder-decoder convolutional neural network to serve as the conditional distribution generator. We call it Fooling Transfer Net (FTN). The details of FTN is described in the next subsection.

\subsection{Network structure} 
Inspired by the image-to-image translation \cite{NIPS2017_6672}, \cite{liu2019few} and style transfer task \cite{gatys2016image}. FTN is built with an encoder $E$ and an AdaIN decoder $D$. We learn the properties of distribution $q(I_{lf})$ from its high-level representations in the victim model, which are described as $\phi_{i}(\hat{I_{lf}})$, where $\hat{I_{lf}}$ is an arbitrary sample sampled from $q(I_{lf})$ and $i$ denotes the targeted layer of the victim model.

The encoder $E$ consists of a sequence of convolutional layers and several residual blocks to encode the source images to a latent vector. The AdaIN decoder is made of three AdaIN Residual Blocks followed by several deconvolutional layers. AdaIN Residual Blocks are the residual blocks with adaptive instance normalization layers, which will first normalize the activations of a sample in each channel to have a zero mean and unit variance and then scale it with learned scalars and biases. In our translation network, the scalars and biases are gotten from the means and variances of $\phi_{i}(\hat{I_{lf}})$. 

Specifically, we extract the $\phi_{i}(\hat{I_{lf}})$ from a pretrained classifier, and put them through two-layer Multilayer Perceptron (MLP) to get a certain number of (depending on the number of encoded features) scalars and biases. These scalars and biases are then used to do the affine transformation to the scaled latent code. Here, we aims at extracting the latent representations of the content from the source images using the encoder and extracting the class-specific representation from fooling image. Then we shift the latent content code using class-specific representation. In this way, we hope to remain the content information of the original images but adjust them to the targeted attribution.

\begin{figure*}
     \centering
         \includegraphics[width=\textwidth]{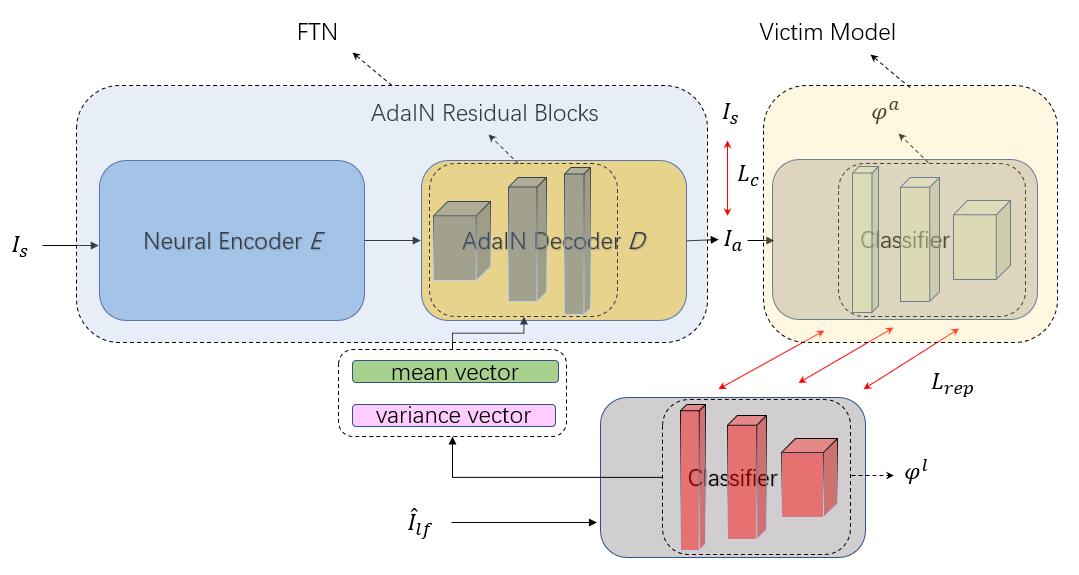}
         \caption{A summary of FTN. Source images $I_{s}$ pass through an encoder-decoder structure to get the final adversarial example $I_{a}$. Sampled low-frequency fooling image $\hat{I}_{lf}$ is sent to a pretrained classifier to get high-level representations $\phi^{l}$. The mean and variance vectors learned from  $\phi^{l}$ are used to scale and shift AdaIN Residual Blocks in the decoder. In the training stage, FTN is supervised by $I_{s}$ in the pixel level, and $\phi^{l}$ in the high-dimensional feature space. }
         \label{fig:FTN}
\end{figure*}

We supervise the network by the original source images $I_{s}$ and high-level representations of the sampled fooling images: $\phi_{i}^{l}$, where $\phi_{i}^{l} = \{\phi_{i}(\hat{I_{lf}}) | \hat{I_{lf}} \sim q(I_{lf})\}\in \mathbb{R}^{d}$. $d = N \times h \times w \times c$, where N is the sampling number, h, w, c, are the height, width, and number of channels of the representation $\phi_{i}(\hat{I_{lf}})$ respectively. $I_{s}$ constrain the output to maintain maximum content information and $\phi_{i}^{l}$ constrain the output to have the similar high-level representations with the fooling images $I_{lf}$. An illustration of FTN is shown in Figure \ref{fig:FTN}. For more details about the network structure, please refer to our code.

\subsection{Loss function} 
We constrain the network by three loss functions, the content loss $L_{c}$, the representation loss $L_{rep}$, and the total variance loss $L_{tv}$. 

The content loss is used to keep content similarity between the adversarial examples and the source images. We use the structural similarity (SSIM) index as our content loss function. SSIM is used to predict the perceived quality of the images. In our experiment, it performs better than the traditional $L_{2}$ loss function. 

The representation loss restrict the output adversarial examples to have similar high-level representations with fooling images in the pretrained classifier. In the paper, we take the high-level representations as the distributions in the high-level space and apply the distribution matching strategy to measure the representations' similarity. For defining a metric of difference between two distributions, we adopt the empirical Maximum Mean Discrepancy (MMD) as the nonparametric metric. MMD is a commonly used metric of discrepancy between two distributions, due to its efficiency in computation and optimization \cite{quadrianto2009distribution}. Denote $\mathbf{\phi}^{l}=\left[\mathbf{\phi}_{1}^{l}, \ldots, \mathbf{\phi}_{n^{l}}^{l}\right] \in \mathbb{R}^{d \times n^{l}}$ and $\mathbf{\phi}^{a}=\left[\mathbf{\phi}_{1}^{a}, \ldots, \mathbf{\phi}_{n^{a}}^{a}\right] \in \mathbb{R}^{d \times n^{a}}$ as the high-level representations of $I_{lf}$ and $I_{a}$ respectively. Denote the data matrix $\mathbf{\phi}=\left[\mathbf{\phi}^{l}, \mathbf{\phi}^{a}\right]$ as the combination of $\phi^{l}$ and $\phi^{a}$. Then MMD between the marginal distributions of the source and the target representations is defined as:
\begin{equation}\label{equation:mmd}
\begin{aligned} \mathcal{L}_{rep} = \mathrm{MMD} &=\left\|\frac{1}{n^{l}} \sum_{i=1}^{n^{l}} \mathrm{\phi}_{i}^{l}-\frac{1}{n^{a}} \sum_{j=1}^{n^{a}} \mathrm{\phi}_{j}^{a}\right\|_{2}^{2} \\ &=\operatorname{Tr}\left(\mathrm{\phi} M \mathbf{\phi}^{T}\right) \end{aligned}
\end{equation}
where M is the MMD matrix. Let $M_{ij}$ be one element of M. $M_{ij}$ can be calculated as:
\begin{equation}\label{equation:mmd_specific}
M_{i j}=\left\{\begin{array}{ll}{1 /\left(n^{l}\right)^{2},} & {i \leqslant l^{l}, j \leqslant n^{a}} \\ {1 /\left(n^{a}\right)^{2},} & {i>n^{l}, j>n^{a}} \\ {-1 /\left(n^{l} n^{a}\right),} & {\text { otherwise }}\end{array}\right.
\end{equation}
A detail of implementation is that we keep the batch size of $I_{a}$ the same as the sampling number of $q(I_{lf})$, so that $\phi^{l}$ and $\phi^{a}$ will have the same dimensions.

We also use the total variance loss to work as a total variation regularization to punish the reconstruction noise.

The total loss of the network is expressed as:
\begin{equation}\label{equation:mse}
\mathcal{L}_{total}=\mathcal{L}_{c} + \gamma \mathcal{L}_{rep} + \lambda \mathcal{L}_{tv} 
\end{equation}
where $\gamma$ and $\lambda$ are the weight constants of $\mathcal{L}_{rep}$ and $\mathcal{L}_{tv}$. The specific setting of these parameters can be referred to our code.
\section{Experiment}
Models:
In the paper, we choose a classic classification model: VGG19 \cite{simonyan2014very} as our \textit{training victim model} and test the transferability on the other more delicate classification models, like Inception-v3 \cite{szegedy2016rethinking}, ResNet-18 \cite{he2016deep}, ResNet-50 \cite{he2016deep} and Densenet \cite{huang2017densely}. We denote them as \textit{validation victim models}. 

Dataset: FTN is trained on ILSVRC 2017 classification training set and tested on its validation set.
 \begin{table*}[h]
\centering
\begin{tabular}{|l | l | l |l |l |l|}
\hline
 & Inception-v3 & Resnet-18 & Resnet-50 & Densenet & Clarifai.com \\
\hline
Niave & 1$\%$ & 2$\%$ & 1$\%$ & 1$\%$ & 0$\%$\\
\hline
TR & 67$\%$ & 79$\%$ & 74$\%$ & 76$\%$ & 51$\%$ \\
\hline
DR & 72$\%$ & 76$\%$&  83$\%$& 75$\%$ & 62$\%$ \\
\hline
TR$\&$DR & 78$\%$ & 81$\%$&  78$\%$& 83$\%$ & 67$\%$ \\
\hline
Gradient-CPPN & 96$\%$ & 94$\%$&  91$\%$& 93$\%$ & 86$\%$ \\
\hline
EA-CPPN & 0$\%$ & 1$\%$ & 0$\%$ & 0$\%$ & 0$\%$\\
\hline
\end{tabular}
\caption{Comparison of fooling images}
\label{tab:fooling image}
\end{table*}
 \begin{figure*}
     \centering
         \includegraphics[width=\textwidth]{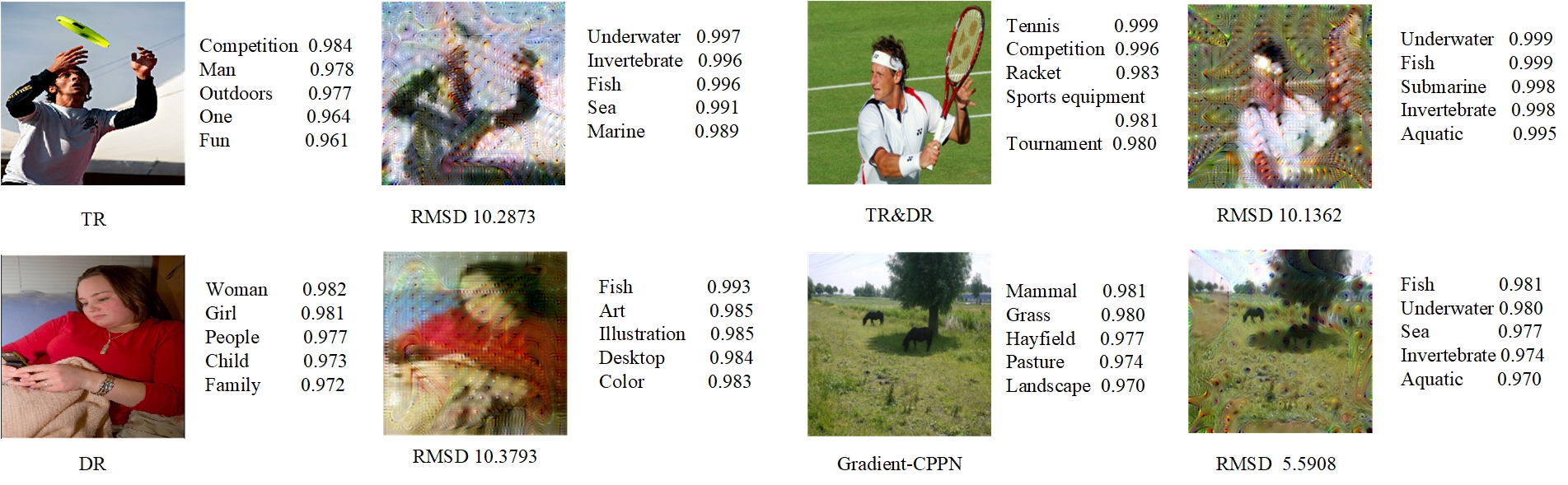}
         \caption{The adversarial examples generated by FTN with different fooling images}
         \label{fig:FTN_dif_fool}
\end{figure*}

Target: We choose attribution 'starfish' as our default targeted class. We choose this class because it is almost impossible to tangle with other classes. The features of starfish are distinct from most of the other objects. Thus it can avoid the situation that the model naturally misclassify the source images as the targeted class and then overestimate the performance of our proposed method. The targeted attack of the other classes can be referred to our project homepage.

Measure: We measure our results by two important factors: transferability and distortion. The transferability is measured by the \textit{transfer success rate}, which means the percentage of the generated adversarial examples are correctly classified as the targeted label by \textit{validation victim models}. The distortion describes the difference between the generated adversarial examples and the source images. We measured the distortion by \textit{Root Mean Square Deviation} (RMSD), which is computed as: $d = \sqrt{\frac{1}{n} \|I_{a}-I_{s}\|_{2}^{2}}$. We also use the Ratio of Transfer success rate and Distortion, which is denoted as RTD, as a measure to compare different methods. It is simply calculated as:
\begin{equation}\label{equation:RTD}
RTD = \frac{transfer\quad success\quad rate}{RMSD} * 100
\end{equation}

\subsection{Low-frequency fooling image}
 In the paper, we proposed low-frequency fooling images and FTN to transfer the source images using low-frequency fooling images. In this section, we aim to prove that $I_{lf}$ are more transferable than $I_{hf}$ and FTN can maintain this transferability. 
 
 In the above, we have given the examples that $I_{lf}$ are more transferable than $I_{hf}$. Here, we do the comprehensive experiment to prove this result. We compare the five high-frequency-constrained methods: CPPN Gradient, CPPN EA, DR, TR, DR+TR and the direct gradient ascent method with no constrain for high-frequency gradients. All of the results are trained on \textit{training victim model} and tested on \textit{validation victim models}. The \textit{transfer success rate} of randomly selected 100 samples are shown in Table \ref{tab:fooling image}. We can see high-frequency constrained methods generally perform better than direct gradient ascent method, and CPPN Gradient method is the best of them. 

 And then, we will show FTN can maintain the transferability of $I_{lf}$. In other words, using more transferable fooling images in FTN can generate more transferable adversarial examples than the others. For the sake of fairness, we adjust the hyper-parameters for every method to get their best effects. RTD and RMSD of them are shown in Table \ref{tab:FTN_self} and the visual comparison is shown in Figure \ref{fig:FTN_dif_fool}. We can see the better class images can contribute to more transferable or less distorted adversarial examples for the same model.

  \begin{table*}[h]
\centering
\begin{tabular}{|l|l | l | l |l |l |l|}
\hline
& RMSD & Inception-v3 & Resnet-18 & Resnet-50 & Densenet & Clarifai.com \\
\hline
TR & 10.13 & 3.94 & 4.34 & 4.65 & 4.21 & 3.00 \\
\hline
DR & 10.28 & 4.52 & 5.33 &  5.43& 5.10 & 3.94 \\
\hline
TR$\&$DR & 10.34 & 6.34 & 6.81&  6.49& 7.02 & 5.41 \\
\hline
Gradient-CPPN & 5.21 & 13.32 & 13.39&  12.13& 12.23 & 12.11 \\
\hline
\end{tabular}
\caption{FTN transferability comparisons using different fooling images}
\label{tab:FTN_self}
\end{table*}
 \begin{table*}[h]
\centering
\begin{tabular}{|l | l | l |l |l |l|l|}
\hline
& RMSD & Inception-v3 & Resnet-18 & Resnet-50 & Densenet & Clarifai.com \\
\hline
FG & 3.56 & 1$\%$ & 2$\%$ & 1$\%$ & 1$\%$ & 0$\%$\\
\hline
JSMA & 3.21 & 2$\%$ & 2$\%$ & 0$\%$ & 1$\%$ & 0$\%$\\
\hline
DeepFool & 3.98 & 28$\%$ & 33$\%$ & 34$\%$ & 31$\%$ & 1$\%$\\
\hline
C$\&$W & 4.55 & 2$\%$ & 3$\%$ & 2$\%$ & 2$\%$ & 0$\%$\\
\hline
FTN & 3.41 & 98$\%$ & 94$\%$ & 93$\%$ & 95$\%$& 94$\%$\\
\hline
\end{tabular}
\caption{Comparison of FTN with different adversarial example generating methods}
\label{tab:FTN_out}
\end{table*}
 \begin{table*}[h]
\centering
\begin{tabular}{|l | l | l |l |l |l|l|}
\hline
& RMSD & Inception-v3 & Resnet-18 & Resnet-50 & Densenet & Clarifai.com \\
\hline
Universal & 16.25 & 63$\%$ & 56$\%$ & 41$\%$ & 51$\%$ & 12$\%$\\
\hline
FTN & 5.68 & 92$\%$ & 88$\%$ & 87$\%$ & 91$\%$& 86$\%$\\
\hline
\end{tabular}
\caption{Comparison of universal attacks}
\label{tab:universal}
\end{table*}
 \subsection{FTN}
 \textbf{AdaIN normalization} We design the AdaIN Residual Blocks in FTN for shifting the latent source distribution by the property of latent class distribution. But in ablation experiment, we see the results are not substantially different. In contrast, the selections of targeted representation layers and the loss functions play more decisive roles. However, we find the AdaIN residual really help the training process to converge more quickly and the setting of hyper-parameters to be more easily. The AdaIN Residual Blocks allow more flexible selection of the decisive hyper-parameter $\gamma$, which denotes the weight of representation loss relative to content loss.
 
\begin{figure*}
     \centering
         \includegraphics[width=\textwidth]{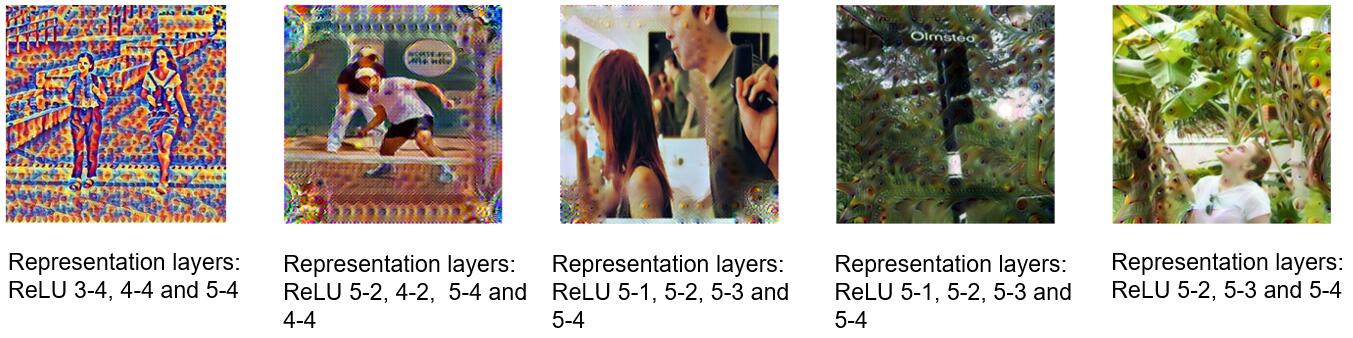}
         \caption{FTN supervised by different layers in VGG19. ReLU 5-4 denotes the fourth ReLU layer in the fifth block in VGG19, and so on. }
         \label{fig:FTN_dif_layer}
\end{figure*}

\textbf{Selection of representations} We have tried many feasible combination of the targeted representation layers in VGG19. In style transfer, the style representations are generally chosen as the activation from low to high. But for generating adversarial examples, the lower representations will endow more superficial similarity between the adversarial examples and the fooling images instead of the semantic features. We choose some typical selections and show them in Figure \ref{fig:FTN_dif_layer}.

More details and parameter settings of training can be referred to our code.
 \subsection{Comparison with other methods}
Different from most of the attacks, we learn the universal mapping in the paper instead of doing optimization for every single image. For the fair comparison, we degrade our method to the single image when comparing with individual attacks. These methods include FG \cite{szegedy2013intriguing}, FGS \cite{goodfellow2014explaining}, Deepfool \cite{moosavi2016deepfool}, JSMA \cite{papernot2016limitations} and C$\&$W attack \cite{carlini2017towards}. We do the optimization on VGG19 model and test on \textit{validation victim models} and Clarifai.com. Clarifai.com is a black-box image classification server. Users can not get access to its dataset, network structure and parameters, which is good for testing the transferability of adversarial examples. As shown in Table \ref{tab:FTN_out}, our method can achieve higher \textit{transfer success rate} than the other methods for the targeted attack.

We also compare our method with another universal attack \cite{moosavi2017universal}. The quantitive results are shown in Table \ref{tab:universal}.
\section{Discussion}
In this paper, we proposed low-frequency adversarial examples and proved it is more transferable than the high-frequency ones. However, what low-frequency attacks mean? Why they can be more transferable than the high-frequency ones? In this section, we attempt to discuss and answer these questions.

Firstly, we find low-frequency attacks and high-freq- uency attacks are highly different. This difference not only lies on the pixels of images but also lies on the classifier representations, which are decisive to the classification (although they will be classified into the same class). This conclusion actually can be confirmed by the observation of the difference between $I_{lf}$ optimized adversarial examples and $I_{hf}$ optimized ones. Because the optimization processes are all the same except the change of class images. But we still do the more rigorous experiments to observe the difference between the representations. We compare the representations of two kinds of fooling images in VGG19 model. Our targeted representation layers are ReLU 5-2, ReLU 5-3, ReLU 5-4, which we find are most effective to supervise our model and most decisive to the final classification. We compute the mean and variance of them and find the mean and variance of the same kind of fooling images are highly similar. This proves although they will be classified as the same class, their high-dimensional features are different.

But what this difference means and how this difference cause the stronger transferability? For answering this question, we analyze the low-frequency attacks and high-frequency attacks on the manifold. In this way, we can think the classification models learn the classified boundaries in a high-dimensional space. The traditional strategy is to learn a high-dimensional vector and add it to a source image point. These methods help the source image to escape from its original boundary (non-targeted attack) or pass trough another specific boundary (targeted attack). We call the vector perturbation. Since these learning methods attempt to find the smallest vector to make the attack successful, these vectors are highly related to classification boundary of victim model. Moosavi \textit{et al}. \cite{moosavi2017universal} took them as the normal vectors from the source image pointing to the boundary. However, although the classification boundaries learned by the different models share some similarities (which make some high-frequency adversarial examples are transferable to some extent), the curvatures of the different boundaries can not be the same. That causes some high-frequency adversarial examples fail to transfer or have to magnify the perturbation to transfer. 

Our method decouples the perturbation from the curvature of the boundary surface. Note that learning a high-frequency fooling image is to find a point that is classified as the targeted class but which may be sensitive to the difference of model boundaries of the different classification. And learning a low-frequency fooling image is to find a point which is still classified as the targeted class but less sensitive to the different model boundaries. We speculate that is because the low-frequency fooling images are closer to the natural image manifold. And the classification boundaries are trained to be more robust to the natural images than the meaningless noise, since the training datasets are consisted of the natural images. In our method, we try to find the point which is near the source image in the pixel space and is close to $I_{lf}$ in the representation space. This constraint condition is independent of the curvatures of the boundaries, thus it is expected to be more transferable than the previous methods. 
 \section{Conclusion}
 In this paper, we first propose a method to produce universal, transferable and targeted adversarial examples. Put specifically, we find constraining high-frequency noises in gradients when attacking a targeted class is able to ensure the transferableability of the fooling images. For developing universal attacks, we then build FTN to learn a universal mapping from source images to adversarial examples. We constrain this mapping by the high-level representations of produced fooling images in a classifier. This ensure the produced adversarial examples are highly transferable, just like the produced fooling images. We test our attacks not only on these widely used classifers, but also on the black-box classification server. The experiments show our attacks are able to mislead the black-box classifiers to our targeted class with very high confidences. When comparing with the other attacks, we attack the classifiers more successfully under more difficult conditions with smaller distortions.
 
 By analyzing the method in the perspective of manifold, we find that our constraint (the similarity of high-level representations of low-frequency fooling images) is robust to the classification boundaries. This may be the reason why our attack is more transferable than the others. 
 
 We think this direction is worth further exploring. For example, one can try to use a neural network discriminator, which is popular recently, to measure the similarity of the high-level representations, or to analyze the fooling image clusters on the manifold, which may further support our hypothesis. We are excited to explore these possibilities in our future works.
\clearpage
{\small
\bibliographystyle{ieee}
\bibliography{egbib}
}

\end{document}